\documentclass{dialogue}

\begin{document}

\begin{otherlanguage}{english}
\begin{center}
{\Large\bfseries{Importance of Copying Mechanism\\for News Headline Generation}}

\medskip

Gusev Ilya (\texttt{ilya.gusev@phystech.edu})

\medskip

MIPT, Moscow, Russia
\end{center}
News headline generation is an essential problem of text summarization because it is constrained, well-defined, and is still hard to solve. Models with a limited vocabulary can not solve it well, as new named entities can appear regularly in the news and these entities often should be in the headline. News articles in morphologically rich languages such as Russian require model modifications due to a large number of possible word forms. This study aims to validate that models with a possibility of copying words from the original article performs better than models without such an option. The proposed model achieves a mean ROUGE score of 23 on the provided test dataset, which is 8 points greater than the result of a similar model without a copying mechanism. Moreover, the resulting model performs better than any known model on the new dataset of Russian news.
\medskip

\end{otherlanguage}

\bigskip

\begin{otherlanguage}{russian}
\begin{center}
{\Large\bfseries{Важность механизма копирования\\для генерации новостных заголовков}}

\medskip

Гусев И. О. (\texttt{ilya.gusev@phystech.edu})

\medskip

МФТИ, Москва, Россия
\end{center}

Генерация заголовков новостей — существенная проблема в области суммаризации (автореферирования) текстов, так как она довольно ограничена, по сравнению с другими типами суммаризации, но всё ещё сложна. Модели с ограниченным словарём будут плохо справляться с такой задачей, потому что новые именованные сущности могут регулярно появляться в новостях, и зачастую они должны быть в заголовках. Для генерации заголовков для новостей на морфологически богатых языках, таких как русский, необходимы модификации моделей из-за обилия возможных словоформ. Цель этой работы — показать, что модели, которые могут копировать слова из оригинальной новости, справляются с задачей генерации заголовков лучше, чем модели без такой возможности. Модель предложенной архитектуры на тестовом наборе данных имеет средний ROUGE, равный 23, что на 8 баллов больше аналогичной модели без возможности копирования. Более того, и на предоставленном тестовом наборе данных, и на наборе данных РИА наша модель показывает результаты лучше, чем какая-либо из известных моделей.
\medskip

\textbf{Ключевые слова:} автореферирование текстов, суммаризация текстов, генерация заголовков, глубокое обучение, seq2seq, copynet, BPE
\end{otherlanguage}

\selectlanguage{english}

\section{Introduction}
Summarization systems are attracting more and more interest due to the increasing number of available texts. One of the main areas of application of these systems is news summarization and headline generation. Good headlines are crucial for news agencies and useful for readers. On the one hand, they should be informative enough and easy to read. On the other hand, they should encourage users to open full articles and should not contain complete information. Different news agencies have a different balance between these two extremes.

The structure of many news articles and especially the ones published online can be described as the inverted pyramid, which means that the first sentence contains critical information and answers basic questions: who, what, where, when, why and how. The rest of the first paragraph includes some important details while the subsequent sections provide more details, background information, and citations. Thereby the first sentence usually contains enough information to be a headline, and it can be a hard baseline to outperform.

Headline generation can be seen as a particular type of abstractive text summarization. Unlike extractive summarization, models of abstractive summarization can generate new words that were not used in the original text. The primary purpose of such models is to capture main information from the original text and produce a shorter version of it.

This work was done within the framework of the headline generation competition of the "Dialogue" conference. Various systems of headline generation were planned to compare within this track. The main training dataset consists of Russian news articles from the ``RIA Novosti'' website\footnote{https://ria.ru} [5]. Another test dataset was closed, but the organizers set up the framework for model evaluation based on Docker.

In addition to these datasets, we used a corpus of news articles of Lenta.Ru for evaluation purposes.

In this paper, we present an approach to headline generation based on two key features over the standard seq2seq model with attention [2]. Byte-pair encoding [9] is the first improvement, and the second is CopyNet [6], which either generates tokens from the vocabulary (as other networks do) or optionally copies a token from the source text. CopyNet model was applied to the task of headline generation in Ayana et al. [1]. We also present the test results of our model on the RIA corpus, Lenta.ru corpus and unknown test dataset from the organizers of the track. These results validate that models with a possibility of copying words from the original article outperform models without such an option.

\section{System description}
The most simple architecture used was encoder-decoder with attention [2] over word-level tokens. One can utilize recurrent neural networks [2] or transformers [13] as encoder and decoder. We tried only LSTM models for encoding and decoding, whereas transformers should perform even better. There are several reasons we did not try to use them. First, there is no default implementation of them as a decoder in the framework we used. Second, they require much time to train on one GPU. Third, Gavrilov et al. [5] utilized Universal Transformer architecture in their work, but we achieved better scores even with LSTM encoder.

The first improvement over simple architecture was byte-pair encoding. It is crucial for many natural language processing tasks in the Russian language as it enables the use of rich morphology and decreases the number of unknown tokens. It often detaches word endings as each word form is less frequent than its stem. Moreover, many words in the Russian language share the same ending, thereby endings of many words can be encoded with the same token. The encoding was trained on the same datasets model trained using sentencepiece [9] library. 

The second improvement was a copying mechanism as described in Gu et al.[6]. Many persons, organizations and locations are typically mentioned in the news. Furthermore, news documents contain unique numbers and dates. Some of these objects should be in the headlines. However, we can not cover all these elements using the fixed size vocabulary. A subword vocabulary can help us in some cases, but it is not enough to deal with this problem entirely. One of the possible solutions to this problem is using the copying mechanism that enables our model to copy tokens from the source text. The most popular solutions are CopyNet [6] and pointer-generator networks [12]. In our system, we used CopyNet primarily because it has an implementation as a part of AllenNLP framework [4].

We used the AllenNLP framework mainly because of its configuration system. Besides, it has a set of necessary modules that can be combined to build a flexible working model. Our code is available online as well as trained models and dataset splits.\footnote{https://github.com/IlyaGusev/summarus}

\section{Data}
\subsection{RIA dataset}
The organizers of the track provided a new dataset of news documents in Russian [5].
This dataset contains texts and titles of around 1 million news document published on the website ``RIA Novosti (RIA news)'' from January 2010 to December 2014. ``Rossiya Segodnya'' (Russia Today) news agency runs this website. Texts were lowercased before publishing. We split the dataset into the train, validation and test parts in a proportion of 90:5:5.

\subsection{Lenta.ru dataset}
Another news dataset available online is the Lenta.ru dataset. It consists of about 800 thousand texts and titles from 1999 to 2018. We utilized this dataset to measure the performance of models trained on the RIA dataset to see how well these models can deal with texts of other style and period.

\subsection{Secret test dataset}
The test dataset of the conference track was available only for evaluation through a Docker container sent to the system. The public leaderboard was available for all participants. The first sentence baseline was hard to beat for this dataset.

We have some assumptions about the structure of this dataset. To begin with, the opening sentences of texts are meaningful. However, that is not true for almost every document in the RIA dataset. First sentences in the RIA dataset contain location, date, and author of the text. Thus either test dataset was sampled from different distribution or was carefully cleaned.

The second observation is linked with evaluation scores obtained. Scores of the models trained on the RIA dataset and evaluated on the Lenta dataset are almost similar with the scores achieved on the secret dataset thus we can consider the hidden dataset to be sampled from a different news agency than RT. The secret dataset could also be from a different time period.

\section{Experiments}

We utilized two architectures, namely standard encoder-decoder model with attention and CopyNet model. Models can have different token types; they can operate on word-level or use byte-pair encoding. We used only LSTM encoder-decoder for reasons described earlier. Moreover, we tried models of different size varying from 5 to 43 million of trainable weights. All models were trained on a single GeForce GTX 1080 with batch sizes depending on the size of the model. We used the loss evaluation on the validation dataset to determine when to stop training. The biggest model was trained for five days.

Byte-pair encoding model was trained on the train part of the RIA dataset. All models have a vocabulary size of 50000 tokens, though there were some experiments with an extended vocabulary. Models that used BPE do not require any additional preprocessing, whereas the texts for word-level models should be tokenized for better performance and the generated titles should be detokenized. The length of the source document was limited to 400 tokens for word-level models and 800 tokens for subword-level models.

No pretrained word or subword embeddings were used in the final models. However, we tried to utilize fastText [3] embeddings for Russian language and pretrained subword embeddings [7], but the performance of the resulting models was not better than the performance of fully trained models.

We did all the training with the Adam optimizer with learning rate 0.001. In most cases, we set a size of the beam search to 10.

The evaluation of models on the secret dataset was done using Docker. Participants were not able to see any error logs of the models. The time limit was set to 30 minutes, so big CopyNet models with beam search could not fit this requirement. We did some batching tricks that enable balancing between time and memory consumption. Eventually, our biggest model can meet the requirements with a beam-search size of 2 and no more.

\section{Results}

We present evaluation results on the RIA datasets in Tab. 1., the secret dataset results in Tab. 2, and the scores of the model trained on the RIA dataset and evaluated on the Lenta dataset in Tab.3. We evaluated our models with the standard ROUGE metric [10], reporting the F1 and recall scores for ROUGE-1, ROUGE-2, and ROUGE-L. ROUGE was measured with a Python package\footnote{https://github.com/bheinzerling/pyrouge}. We used the mean of ROUGE-1-F, ROUGE-2-F, and ROUGE-L-F as the primary metric (R-mean-f).  Also, we utilized BLEU [11]  as an internal metric as it was easier to measure before we found a decent Python package for measuring ROUGE.
\begin{table}[htbp]
\centering
\begin{tabular}{|c||c|c|c|c|c|c||c|c|}\hline
Model & R-1-f & R-1-r & R-2-f&R-2-r&R-L-f&R-L-r&R-mean-f&BLEU\\\hline\hline
seq2seq-bpe-5m & 38.78 & 36.91 & 21.87 & 20.90 & 35.96 & 35.24 & 32.20 & 49.77 \\\hline
copynet-words-10m & 39.48 & 38.39 & 22.57 & 22.05 & 36.95 & 36.69 & 33.00 & 51.99 \\\hline
copynet-bpe-10m & 40.03 & 38.68 & 23.25 & 22.50 & 37.44 & 37.04 & 33.57 & 52.57 \\\hline
seq2seq-words-25m & 36.96 & 35.19 & 19.68 & 19.02 & 34.30 & 33.60 & 30.31 & 44.69\\\hline
seq2seq-bpe-25m & 40.30 & 38.83 & 22.94 & 22.18 & 37.50 & 37.01 & 33.58 & 51.66 \\\hline
copynet-words-25m & 40.38 & 39.46 & 23.26 & 22.83 & 37.80 & 37.70 & 33.81 & 52.99 \\\hline
copynet-bpe-43m & \textbf{41.61} & 40.33 & \textbf{24.46} & \textbf{23.76} & \textbf{38.85} & 38.51 & \textbf{34.97} & \textbf{53.80} \\\hline\hline
First Sentence [5] & 24.08 & \textbf{45.58} & 10.57 & 21.30 & 16.70 & \textbf{41.67} & 17.12 & - \\\hline
UTransformer [5] & 39.75 & 37.62 & 22.15 & 21.04 & 36.81 & 35.91 & 32.90 & - \\\hline
\end{tabular}
\caption{RIA dataset evaluation}
\end{table}

The baseline provided by the organizers was reasonably straightforward. They split a text into sentences and used the first one as the title. We slightly modified this baseline to achieve a better score. We removed full stops and constrained the number of words used as the title to 25. These steps seem reasonable as full stops are rarely used in news titles and titles should not be too long.

One can see that models with subword tokens perform better than the ones with word tokens of the same trainable weights count and vocabulary size. For example, R-mean-f for the word-level encoder-decoder model with 25 million weights is 30.31, which is significantly lower than R-mean-f for the subword-level model of the same weights count. 

\begin{table}[htbp]
\centering
\begin{tabular}{|c|c|}\hline
Model & R-mean-f\\\hline\hline
seq2seq-bpe-5m & 14.85\\\hline
seq2seq-bpe-25m & 15.40  \\\hline
copynet-words-10m & 20.49* \\\hline
copynet-bpe-10m & 21.69  \\\hline
copynet-bpe-43m & \textbf{23.00} \\\hline\hline
First Sentence & 19.50 \\\hline
First Sentence (modified) & 19.89 \\\hline
\end{tabular}
\caption{Secret dataset evaluation. The score of * was without detokenization.}
\end{table}

The other observation is CopyNet having scores higher than standard encoder-decoder models with attention. That is especially true for the secret dataset where there were no successful tries to beat the first sentence baseline without the copying mechanism. Moreover, as we know, no other participants succeeded to outperform this baseline.

We did an ablation study for BPE and copying mechanism with models of 25 million trainable weights. Copying mechanism has a more significant impact than BPE, improving R-mean by 3.42 R-mean-f points on the RIA dataset, whereas BPE adds 3.27 R-mean-f points.

\begin{table}[htbp]
\centering
\begin{tabular}{|c||c|c|c|c|c|c||c|c|}\hline
Model & R-1-f & R-1-r & R-2-f&R-2-r&R-L-f&R-L-r&R-mean-f&BLEU\\\hline\hline
seq2seq-bpe-5m & 19.38 & 17.35 & 8.27 & 7.43 & 16.94 & 16.55 & 14.86 & 25.14 \\\hline
seq2seq-words-25m & 18.29 & 17.11 & 7.21 & 6.96 & 16.23 & 16.13 & 13.91 & 23.35\\\hline
seq2seq-bpe-25m & 20.75 & 19.06 & 8.77 & 8.11 & 18.15 & 17.97 & 15.89 & 28.21 \\\hline
copynet-words-25m & 28.24 & 27.51 & 13.67 & 13.51 & 25.67 & 25.91 & 22.53  & 40.13\\\hline
copynet-words-10m & 26.37 & 26.38 & 12.67 & 12.74 & 24.04 & 25.06 & 21.02 & 38.36 \\\hline
copynet-bpe-10m & 25.60 & 24.57 & 12.33 & 11.84 & 23.03 & 23.33 & 20.32 & 36.13\\\hline
copynet-bpe-43m & \textbf{28.27} & 27.61 & \textbf{13.95} & 13.63 & \textbf{25.77} & 26.19 & \textbf{22.66} & \textbf{40.44} \\\hline\hline
First sentence & 25.45 & \textbf{40.52} & 11.16 & \textbf{18.63} & 19.17 & \textbf{37.80} & 18.59 & 25.45\\\hline
\end{tabular}
\caption{Lenta dataset evaluation with a model trained on RIA dataset}
\end{table}

We also measured the scores of models trained on the RIA dataset applied to the Lenta dataset. News agencies have their own writing style of headlines, so we need to validate that models capture the essence of an article and not that style. Moreover, texts of RIA and Lenta articles also differ, so it should be harder for models to condition on Lenta articles. These scores confirm that copying mechanism is essential when transferring models between different news agencies. Remarkably, in some cases, BPE worsens models with copying mechanism. One of these cases was on the Lenta dataset with two models of 10 million weights.

\begin{table}[htbp]
\centering
\begin{tabular}{|c|c|c|}\hline
Model & Lenta & Secret \\\hline\hline
seq2seq-bpe-5m & 14.86 & 14.85 \\\hline
seq2seq-bpe-25m & 15.89 & 15.40 \\\hline
copynet-bpe-10m & 20.32 & 21.69  \\\hline
copynet-bpe-43m & 22.66 & 23.00 \\\hline
First sentence & 18.59 & 19.50 \\\hline
\end{tabular}
\caption{R-mean-f scores: Lenta dataset vs Secret dataset}
\end{table}

In Tab. 4 we provide a comparison between scores achieved on the secret dataset and scores achieved on the Lenta dataset. They are almost similar. We suppose that the organizers of the competition used some part of the Lenta dataset for evaluation as the secret dataset. We can explain slight differences in scores with another dataset split. The Lenta dataset contains period including one from  2010 to 2014 (period of the RIA dataset), so different dataset splits can contain this interval or not, and it can dramatically influence scores.

\section{System and error analysis}

We provide three examples of bad cases for most models. We do not state texts of articles here, but they are available in the RIA dataset. However, we have a reference title to catch what is this article about.

Almost in all severe cases, the model without both BPE or copying mechanism produces UNK tokens. Headlines with this token will be considered wrong by users. These type of errors are worse than wrong word forms sometimes produced by subword models.

\begin{table}[htbp]
\centering
\begin{tabular}{|c|p{110mm}|}\hline
Reference title & дело в отношении бельтюкова не скажется на "сколково" - вексельберг\\\hline\hline
seq2seq-words-25m & "сколково" UNK возбуждение дела против UNK\\\hline
seq2seq-bpe-5m & "сколково" обеспокоен возбуждением дела против экс-главы фонда \\\hline
seq2seq-bpe-25m & "сколково" считает возбуждение дела против вице-президента\\\hline
copynet-words-10m & ситуация против бельтюкова не скажется на воплощении проекта " сколково " \\\hline
copynet-bpe-10m &  "сколково" озабочено возбуждение дела против бельтюкова \\\hline
copynet-bpe-43m & руководство "сколково" озабочено возбуждение дела против бельтюкова \\\hline
\end{tabular}
\caption{Beltukov example}
\end{table}

We have a person named Beltukov (бельтюков) in the example in Tab. 5. In the case of the word-level model without copying mechanism, he appears in the headline as UNK.  He can appear in the headline as Beltukov only with the usage of the source tokens. 

\begin{table}[htbp]
\centering
\begin{tabular}{|c|p{110mm}|}\hline
Reference title & "бенфика" и "севилья" сыграют в турине в финале футбольной лиги европы\\\hline\hline
seq2seq-words-25m & "бенфика" и "UNK" сыграют в финале лиги европы\\\hline
seq2seq-bpe-5m & "бенфика" и "севильи" сыграют в финале лиги европы\\\hline
seq2seq-bpe-25m & футболисты "бенфики" и "севильи" сыграют в финале лиги европы\\\hline
copynet-words-10m & "бенфика" и "севилья" сыграют в финале лиги европы\\\hline
copynet-bpe-10m & "бенфика" и "севилья" сыграют в финале лиги европы\\\hline
copynet-bpe-43m & "бенфика" и "севилья" сыграют в финале лиги европы\\\hline
\end{tabular}
\caption{Sevilla example}
\end{table}

There are two football clubs in the example in Tab. 6: Benfica (бенфика) and  Sevilla (севилья). Benfica is in the vocabulary of the word-level model, whereas Sevilla is not. Small subword-level model without copying mechanism tried to reconstruct the name of the club but made an error ("севильи" instead of "севилья"). All the CopyNet models generated decent headlines.

\begin{table}[htbp]
\centering
\begin{tabular}{|c|p{110mm}|}\hline
Reference title & лучшей смерти он себе и не желал – вдова журналиста марка дейча\\\hline\hline
seq2seq-words-25m & UNK\\\hline
seq2seq-bpe-5m & день на бали\\\hline
seq2seq-bpe-25m & марк дейча\\\hline
copynet-words-10m & "за заслуги перед отечеством": "за заслуги перед отечеством" \\\hline
copynet-bpe-10m & друзья и родные проводили в последний путь журналиста марка дейча\\\hline
copynet-bpe-43m & марк дейч утонул на острове бали\\\hline
\end{tabular}
\caption{Mark Deutch example}
\end{table}

The example in Tab.7 was tough for some models. The death of Mark Deutch ("марк дейч") was described in this article. Word-level models failed to generate a headline for this example. Subword word-level models caught whether the place or the person. CopyNet models with subwords succeeded to generate a meaningful headline. 

\section{Conclusion and future work}
It was surprising not to see any other participants successfully using the CopyNet or pointer-generator networks, as these techniques are commonly used for other tasks of the text summarization.

In conclusion, our results validate that the copying mechanism applies to the task of headline generation well. Moreover, in most cases, it is required to use this mechanism due to a variety of named entities, numbers and dates in the news.

As for future work, we should try the transformer as a decoder, make an evaluation of pointer-generator networks for this task, and utilize reinforcement learning methods described in Keneshloo et al. [8]. Besides, it would be nice to make a human evaluation and extend these models to other languages.

\subsubsection*{Acknowledgements}

Authors are thankful to Kozlovskiy Borislav, Lyubanenko Vadim and Smirnova Elizaveta for proofreading and to Smurov Ivan for useful advises.

\color{blue}\section*{References}

\makeatletter
\renewcommand{\section}{\@gobbletwo}
\makeatother
\begin{enumerate}
    \item Ayana, Shen S., Lin Y. et al. (2017), Recent advances on neural headline generation, Journal of computer science and technology. pp. 768–784.
    \item Bahdanau D., Cho K., Bengio Y. (2014), Neural Machine Translation by Jointly Learning to Align and Translate, arXiv:1409.0473.
    \item Bojanowski P., Grave E., Joulin A., Mikolov T. (2017), Enriching Word Vectors with Subword Information, Transactions of the Association for Computational Linguistics, pp. 135-146.
    \item Gardner M, Grus J. et al. (2017), AllenNLP: A Deep Semantic Natural Language Processing Platform, Proceedings of Workshop for NLP Open Source Software. pp. 1-6.
    \item Gavrilov D., Kalaidin P., Malykh V. (2019), Self-Attentive Model for Headline Generation, arXiv:1901.07786.
    \item Gu J., Lu Z., Li H., Li V. (2016), Incorporating Copying Mechanism in Sequence-to-Sequence Learning, Proceedings of the 54th Annual Meeting of the Association for Computational Linguistics (Volume 1: Long Papers), pp. 1631–1640.
    \item Heinzerling B., Strube M. (2018), BPEmb: Tokenization-free Pre-trained Subword Embeddings in 275 Languages, Proceedings of the Eleventh International Conference on Language Resources and Evaluation.
    \item Keneshloo Y., Ramakrishnan N., Reddy C. (2018), Deep Transfer Reinforcement Learning for Text Summarization, arXiv:1810.06667.
    \item Kudo T., Richardson J. (2018), SentencePiece: A simple and language independent subword tokenizer and detokenizer for Neural Text Processing, Proceedings of the 2018 Conference on Empirical Methods in Natural Language Processing: System Demonstrations, pp. 66–71.
    \item Lin C.Y. (2004), ROUGE: A package for automatic evaluation of summaries, Text summarization branches out: ACL workshop.
    \item Papineni, K., Roukos, S., Ward, T., Zhu, W. J. (2002), BLEU: a method for automatic evaluation of machine translation, 40th Annual meeting of the Association for Computational Linguistics, pp. 311–318.
    \item See A., Liu J. P., Manning C. (2017), Get To The Point: Summarization with Pointer-Generator Networks, Proceedings of the 55th Annual Meeting of the Association for Computational Linguistics (Volume 1: Long Papers), pp. 1073–1083.
    \item Vaswani, A., Shazeer, N., Parmar, N., Uszkoreit, J., Jones, L., Gomez, A.N., Kaiser, L., Polosukhin, I. (2017), Attention Is All You Need, Advances in Neural Information Processing Systems, pp. 5998–6008.
    
\end{enumerate}

\end{document}